\documentclass[10pt]{article}

\usepackage{amsmath,amsbsy,amssymb,amsthm,amsfonts}
\usepackage{multirow,multicol,booktabs,wrapfig}
\usepackage{graphicx,float,stfloats,subcaption}
\usepackage{algorithm,algorithmic}
\usepackage[table]{xcolor}
\PassOptionsToPackage{hyphens}{url}
\usepackage{hyperref}
\usepackage[a4paper, margin=0.5in]{geometry}

\title{Quality Inference \\ in Federated Learning with Secure Aggregation}

\author{Bal{\'a}zs Pej{\'o} and Gergely Bicz{\'o}k}
\date{CrySyS Lab, Department of Networked Systems and Services, \\
	Faculty of Electrical Engineering and Informatics,\\
	Budapest University of Technology and Economics,\\ 
		\{pejo,biczok\}@crysys.hu}

\begin{document}

\maketitle

\textbf{Abstract.} Federated learning algorithms are developed both for efficiency reasons and to ensure the privacy and confidentiality of personal and business data, respectively. Despite no data being shared explicitly, recent studies showed that the mechanism could still leak sensitive information. Hence, secure aggregation is utilized in many real-world scenarios to prevent attribution to specific participants. In this paper, we focus on the quality (i.e., the ratio of correct labels) of individual training datasets and show that such quality information could be inferred and attributed to specific participants even when secure aggregation is applied. Specifically, through a series of image recognition experiments, we infer the relative quality ordering of participants. Moreover, we apply the inferred quality information to stabilize training performance, measure the individual contribution of participants, and detect misbehavior.

\section{Introduction}
\label{sec:intro}

For machine learning (ML) tasks, it is widely accepted that more training data leads to a more accurate model. Unfortunately, in reality, the data is scattered among multiple different entities. Thus, data holders could potentially increase the accuracy of their local model accuracy by training a joint model together with others \cite{pejo2019together}. Several collaborative learning approaches were proposed in the literature, amongst which the least privacy-friendly method is centralized learning, where a server pools the data from all participants together and trains the desired model. On the other end of the privacy spectrum, there are cryptographic techniques such as multi-party computation \cite{cramer2015secure} and homomorphic encryption \cite{gentry2009fully}, guaranteeing that only the final model is revealed to legitimate collaborators and nothing more. Neither of these extremes admits most real-world use cases: while the first requires participants to share their datasets directly, the latter requires too much computational resource to be a practical solution for big data scenarios. 

Somewhere between these (in terms of privacy protection) stands \emph{federated learning} (FL), which mitigates the communication bottleneck and provides flexible participation by selecting a random subset of participants per round, who compute and send their model updates to the aggregator server \cite{konecny_federated_2016}. FL provides some privacy protection by design as the actual data never leaves the hardware located within the participants' premises. Yet, there is already rich and growing related literature revealing that from these updates (i.e., gradients) a handful of characteristics can be inferred about the underlying training dataset. Potential attacks include model inversion \cite{fredrikson2015model}, membership inference \cite{shokri2017membership}, reconstruction attack \cite{zhu2019deep}, (hyper)parameter inference \cite{tramer2016stealing}, and property inference \cite{melis2019exploiting}. 

Parallel to these, several techniques have been developed to conceal the participants' updates from the aggregator server, such as differential privacy (DP) \cite{desfontaines2020sok} and \emph{secure aggregation} (SA) \cite{mcmahan2016communication}. Although DP comes with a mathematical privacy guarantee, it also results in heavy utility loss, which limits its applicability in many real-world scenarios. On the other hand, \emph{SA} does not affect the aggregated final model, which makes it a suitable candidate for many applications. Essentially, \emph{SA} hides the individual model updates without changing the aggregated model by adding pairwise masks to the participants' gradients in a clever way so that they cancel out during aggregation. 

Consequently, \emph{SA} only protects the participants' individual updates and leaves the aggregated model unprotected. Hence, \emph{SA} provides a ``hiding in the crowd'' type of protection \cite{sweeney2002k}, thus, without specific background knowledge, it is unlikely that a privacy attacker could link the leaked information to a specific participant. The lack of attribution severely affects the security of FL as well; we are not aware of any attack detection scheme applicable with \emph{SA} enabled.

In this paper, we study the possibility of \textit{inferring the quality of the individual datasets when SA is in place}. This could also be utilized for attack detection as well. Note, however, that it is different from mere poisoning and backdoor detection \cite{bagdasaryan2020backdoor}, as that line of research is only interested in classifying participants as malicious or benign, while our goal is to enable the fine-grained differentiation of FL participants with respect to their data quality. This is fundamentally similar to contribution score computation, which is also an unsolved problem in the \emph{SA} setting. 

Data quality is a complex concept with multiple dimensions \cite{batini2016data}. Moreover, it is relative: it can only be considered in terms of the proposed use, and in relation to other data samples. For this reason (similarly to \cite{wang2020principled}) we focus on image recognition tasks with noisy labels, as in this scenario data quality has a straightforward interpretation. 

\subsection*{Contributions}

We propose a method called \emph{Quality Inference} (QI) which (by utilizing the improvement of the aggregated updates) recovers the relative label quality of the contributing participants' datasets. To obtain this quality information, our method takes advantage of the improvements of the aggregated models across multiple rounds, as well as the known per-round selected subset of participants. QI works by evaluating the aggregated updates in each round and assigning scores to the selected participants based on three simple but novel rules called \emph{The Good}, \emph{The Bad}, and \emph{The Ugly} (as in the movie \cite{movie}). As a result, we are able to recover the relative quality ordering (i.e., by label correctness rate) of the participants.

We simulated datasets with different qualities by utilizing unique label-flipping rates for each participant, and conduct experiments on two neural network architectures (MLP and CNN) and two datasets (MNIST and CIFAR10). We consider three FL settings, where 2 out of 5, 5 out of 25, and 10 out of 100 participants are selected in each round to update the model, respectively. 

Our experiments show that the three proposed heuristic scoring rules significantly outperform the baseline in determining the participants' data qualities relative to each other (i.e., correct label rates). We find that the accuracy of QI depends on both the complexity of the task and the trained model architecture. We also conduct an ablation study on the hyperparameters of the proposed rules. 

Finally, we investigate three potential applications of QI: on-the-fly performance boosting, contribution score computation, and misbehavior detection (by considering free-riding and poisoning). We find that i) carefully weighting the participants based on the inferred scores smooths the learning curve, ii) the scores could be used as a measure of participant contribution, and iii) the scores are able to reveal misbehaving participants. This latter implies that besides the label correctness rate, QI is also capable of inferring other, more general quality aspects of the data. We are not aware of any work tackling any of the aforementioned issues when \emph{SA} is enabled.
\section{The Theoretical Model}
\label{sec:model}

In this section, we introduce the theoretical model of quality inference and highlight its  complexity. We note with $n$ a participant in FL, while $N$ denotes the number of all participants. Similarly, $i$ denotes a round in FL, while $I$ denotes the number of all rounds. The set $S_i$ contains the randomly selected participants for round $i$, and $b=|S_i|$ captures the number of selected participants. $D_n$ is participant $n$'s dataset consisting of $(x,y)\in D_n$ data-label pairs. We assume $D_n$ is associated with a single scalar $u_n$, which measures its quality. We use $\theta_n$ and $v_i$ to capture the quality of the $n$th participant's gradient and the quality of the aggregated gradient in the $i$th round, respectively. A summary of the variables is listed in the Appendix (Table \ref{tab:variables}).

\subsection{Deterministic Case}

In this simplified scenario, we assume the gradient quality is equal to the dataset quality, i.e.,  $\theta_n=u_n$. Consequently, the aggregated gradients represent the average quality of the participants' datasets. As a result, the round-wise quality values of aggregated gradients form a linear equation system $Au=v$, where $u=[u_1,\dots,u_N]^T$, $v=[v_1, \dots, v_I]^T$, and $a_{i,n}\in A_{I\times N}$ indicates whether participant $n$ is selected for round $i$. Depending on the dimensions of $A$, the system can be under- or overdetermined. In case of $I<N$ (i.e., no exact solution exists) and if $I>N$ (i.e., many exact solutions exist), the problem itself and the approximate solution are shown in Eq. \ref{eq:over} and \ref{eq:under}, respectively. 

\begin{align}
	\min_u||v-Au||_2^2
	\hspace{0.5cm}\Rightarrow\hspace{0.5cm}
	u=(A^TA)^{-1}A^Tv \label{eq:over}\\
    \min_u||u||_2^2 \text{ s.t. } Au=v
	\hspace{0.25cm}\Rightarrow\hspace{0.25cm}
	u=A^T(AA^T)^{-1}v \label{eq:under}
\end{align}

\subsection{Stochastic Case}

The above equations do not take into account any randomness. Given that the training is stochastic, we can treat the quality of participant $n$'s gradient as a random variable $\theta_n$ sampled from a distribution with parameter $u_n$. Moreover, we can represent $\theta_n=u_n+e_n$ where $e_n$ corresponds to a random variable sampled from a distribution with zero mean. We can further assume that $e_n$ and $e_{n'}$ are i.i.d. for $n \ne n'$. As a result, we can express the aggregated gradient $v_i=\sum_n a_{i,n}u_n+E$ where $E$ is sampled from the convolution of the probability density function of $e$'s. 

In this case, due to the Gauss–Markov theorem \cite{harville1976extension}, the solution in Eq. \ref{eq:over} is the best linear unbiased estimator, with error $||v-Au||_2^2=v^T(\textbf{I}-A(A^TA)^{-1}A^T)v$ (where $\textbf{I}$ is the identity matrix) with an expected value of $b(\textbf{I}-N)$. 
Note, that with more iterations more information is leaking, which should decrease the error. Yet, this is not captured by the theorem as it considers every round as a new constraint.

This problem lies within estimation theory \cite{ludeman2003random}, from which we already know that estimating a single random variable with added noise is already hard; moreso, factoring in that in our setting, we have multiple variables forming an equation system. Moreover, these random variables are different per round; a detail we have omitted thus far. Nevertheless, each iteration corresponds to a different expected accuracy improvement level, as with time the iterations improve less and less. Consequently, to estimate individual dataset quality we have to know the baseline expected learning curve; in turn, the learning curve depends exactly on those quality values. Being a chicken-egg problem, we focus on empirical observations to break this vicious cycle.
\section{Quality Scoring}
\label{sec:QI}

In this section we devise the three intuitive scoring rules which are the core of QI: they either reward or punish the participants in the FL rounds. The notations used in this section are summarized in Table \ref{tab:variables1} and \ref{tab:variables2}. We define $\omega_i$ as the aggregated model's improvement in the $i$th round and $\varphi_{i,n}$ as the quality score of participant $n$ after round $i$. Note that in the rest of the paper we slightly abuse the notation by removing index $i$ where it is not relevant. 

\begin{table}[t]
	\centering
	\parbox{.45\linewidth}{
	\begin{tabular}{c|p{5.5cm}}
		Variable & Description \\
		\midrule
		$n\in [N]$ & Participants \\
		$i\in [I]$ & Training rounds\\
		$S_i$ & Set of selected par. for round $i$\\
		$b$ & Number of selected participants \\
		$(x,y)\in D_n$ & Dataset of participant $n$\\
		$u_n$ & Quality of $D_n$\\
		$v_i$ &  Quality of aggr. gradient in round $i$\\
		$\theta_n$ & Quality of participant $n$'s gradient\\
	\end{tabular}
	\caption{Notation used in FL.}
	\label{tab:variables}}
	\hfill
	\parbox{.45\linewidth}{
	\begin{tabular}{c|p{6cm}}
		Variable & Description \\
		\midrule
		$\omega_i$ & Model improvement in the $i$th round \\
		$\varphi_{i,n}$ & Quality score of par. $n$  after round $i$\\
		$q_{i,n}$ & Inferred quality-wise rank of participant $n$ after round $i$\\
		$d_s$ & Spearman Distance\\
		$r_s$ & Spearman Coefficient\\
	\end{tabular}
	\caption{Notation used in QI.}
	\label{tab:variables2}}
\end{table}

\subsection{Assumptions}

We assume an honest-but-curious setting; the aggregator server (and the participants) cannot deviate from the FL protocol. Further restrictions on the attacker include limited computational power and no background knowledge besides access to an evaluation oracle. For this reason, we neither utilize any contribution score based techniques nor existing inference attacks, as these require either significant computational resources or user-specific relevant background information.

\subsection{Scoring Rules}

Based on the round-wise improvements $\omega_i$, we created three simple rules to reward or punish the participants. We named them \emph{The Good}, \emph{The Bad}, and \emph{The Ugly} (as in the spaghetti western movie \cite{movie}); the first one (G) rewards the participants in the more useful aggregates, the second one (B) punishes in the less useful ones, while the last one (U) punishes when the aggregate does not improve the model at all. 

\begin{itemize}
	\item[G] Each participant $n^\prime$ contributing in round $i$ that improves the model more than the previous round (i.e., $\omega_i>\omega_{i-1}$) receives $+1$, i.e., $\varphi_{i,n^\prime} \leftarrow \varphi_{i-1,n^\prime} + 1$.
	\item[B] Each participant $n^\prime$ contributing in round $i$ that improves the model less than the following round (i.e., $\omega_i<\omega_{i+1}$) receives $-1$, i.e., $\varphi_{i+1,n^\prime} \leftarrow \varphi_{i,n^\prime} - 1$.
    \item[U] Each participant $n^\prime$ contributing in round $i$ that does not improve the model at all (i.e., $\omega_i<0$) receives $-1$, i.e., $\varphi_{i,n^\prime} \leftarrow \varphi_{i-1,n^\prime} - 1$.
\end{itemize}

Note, that the quality score in round $i$ is only updated for participant who has contributed in that round (\emph{The Good} and \emph{The Ugly}) and the previous round (\emph{The Bad}). 
For instance, if in round $i$ the improvement was negative and in the following round it was positive, the participants of round $i$ receive $-1$ due to \emph{The Ugly} in round $i$ and receive another $-1$ in round $i+1$ due to \emph{The Bad}. 
For the rest of the participants (noted with $\hat{n}$), the scores remain unchanged, i.e., $\varphi_{i,\hat{n}} \leftarrow \varphi_{i-1,\hat{n}}$. 

It is reasonable to expect that the improvements in consecutive rounds are decreasing (i.e., $\omega_i<\omega_{i-1}$): first the model improves rapidly, while improvement slows down considerably in later rounds. The first two scoring rules (\emph{The Good} and \emph{The Bad}) capture the deviation from this pattern: we can postulate that i) high dataset quality increases the improvement more than in the previous round, and ii) low dataset quality decreases the improvement, which would be compensated in the following round. These phenomena were also shown in \cite{kerkouche2020federated}. While these rules are relative, the last one (\emph{The Ugly}) is absolute: it builds on the premise that if a particular round does not improve the model, there is a higher chance that some of the corresponding participants have supplied low-quality data.

Independently of the participants' dataset qualities, round-wise improvements could deviate from this pattern owing to the stochastic nature of learning. We postulate that this affects all participants evenly, independently of their dataset quality; thus, the ordering among the individual scores is not significantly affected by this ``noise''. Participant selection also introduces a similar effect; however, we assume that participants are selected uniformly, hence, its effect should also be similar across participants. 
\section{Experimental Setup}
\label{sec:setup}

In this section, we describe our experimental setup, including the evaluation metric, the quality simulation, and the utilized datasets and model architectures.

\subsection{Evaluation Metric}

The quality scores of the participants are unlikely to converge; hence, we focus on their ordering. We denote with $q_{i,n}$ the inferred quality-wise rank of participant $n$ after round $i$, and we measure the accuracy of the inferred qualities by comparing $q_{i,n}$ for each participant to the baseline quality-wise ordering. For this purpose, we use the Spearman correlation coefficient $r_s$ \cite{zar2005spearman}, which is based on the Spearman distance $d_s$ \cite{diaconis1977spearman} (as seen in Eq. \ref{eq:qi}). The Spearman distance measures the absolute difference between this inferred and the actual position, while the Spearman correlation coefficient assesses monotonic relationships on the scale $[-1,1]$; $1$ corresponds to perfect correlation, while any positive value signals a positive correlation between the actual and the inferred quality ordering. E.g., if the inferred quality order (via the three rules) expressed with participant IDs is 5-3-2-4-1, while the actual quality order is 5-4-3-2-1, then the Spearman distances are 0-2-1-1-0, and the Spearman correlation is 0.7, suggesting that the inferred quality order is very close to the original one. Note, that the Spearman distance (and consequently the coefficient) handles any misalignment equally, irrespective of the position. 

\begin{equation}
	\label{eq:qi}
	d_s(i,n) = |n-q_{i,n}|
	\hspace{0.25cm}
	r_s(i)=1-\frac{6\cdot \sum_{n=1}^Nd_s(i,n)^2}{N\cdot(N^2-1)}
\end{equation}

\subsection{Simulating Data Quality}

Data quality can only be considered in terms of the proposed use and in relation to other data samples, i.e., participants with different data distributions could have different views of the same dataset. To tackle this issue, we consider only the IID case in our experiments. Besides, data quality entails multiple aspects such as accuracy, completeness, redundancy, readability, accessibility, consistency, usefulness, and trust, with several having their own subcategories \cite{batini2016data}. In this paper, we focus on image recognition tasks as it is a key ML task with standard datasets available. Still, we have to consider several of these aspects in relation to image data. 

Unfortunately, we are not aware of any public datasets encompassing data from several well-categorized quality classes. Since visual perception is a complex process, to avoid serious pitfalls, we do not manipulate the images themselves, but simulate different qualities similarly to \cite{wang2020principled}: we modify the label $y$ corresponding to a specific image $x$. To have a clear quality-wise ordering between the datasets (i.e., the ground truth), we perturbed the labels of the participants according to Eq. \ref{eq:scramble}, where $\psi_k$ is drawn uniformly at random over all available labels. Putting it differently, the labels of the participants' datasets are randomized before training with a linearly decreasing probability, e.g., in the case of five participants with IDs [1,2,3,4,5], the ratio of assigned random labels are 100\%, 75\%, 50\%, 25\%, and 0\%, respectively.

\begin{equation}
	\label{eq:scramble}
    \Pr(y_k=\psi_k|(x_k,y_k)\in D_n)=\frac{N-n}{N-1}
\end{equation}

\subsection{Datasets, ML Models and Experiment Setup}

For our experiments, we used the MNIST \cite{deng2012mnist} and the CIFAR10 \cite{krizhevsky2014cifar} datasets. MNIST corresponds to the simple task of digit recognition. It contains $70,000$ hand-written digits in the form of $28 \times 28$ gray-scale images. CIFAR10 is more involved, as it consists of $60,000$ $32 \times 32$ color images of various objects. For MLP, we used a three-layered structure with hidden layer size 64, while for CNN, we used two convolutional layers with 10 and 20 kernels of size 5$\times$5, followed by two fully-connected hidden layers of sizes 120 and 84. For the optimizer, we used SGD with a learning rate of 0.01 and a dropout rate of 0.5. The combination of the two datasets and the two neural network models yields four use cases. In the rest of the paper, we will refer to these as MM for MLP-MNIST, MC for MLP-CIFAR10, CM for CNN-MNIST, and CC for CNN-CIFAR10. 

We ran all the experiments for 100 rounds and with three different FL settings, corresponding to 5, 25, and 100 participants where 2, 5, and 10 of them are selected in each round, respectively. The three FL settings combined with the four use cases result in twelve evaluation scenarios. We ran every experiment 10-fold, with randomly selected participants. 

\subsection{Empirical Quality Scores}

\begin{algorithm}[b!]
    \small
	\caption{QI in FL with SA}
	\label{alg:fedscore}
    \textbf{Input}: data $D$; participants $N$; rounds $I$
    
	\begin{algorithmic}[1]
	    \STATE $\mathrm{Split}(D,N)\rightarrow \{D_1,\dots,D_N,D_{N+1}\}$
	    \STATE \textbf{for }$n\in [1,\dots,N]$\textbf{ do }
	        \STATE \hspace{0.5cm} $\forall$ $(x_k,y_k)\in D_n: y_k\sim$ Eq. \ref{eq:scramble}
		\STATE $\varphi=[0,\dots,0]$; $M_0\leftarrow \mathrm{Rand}()$
		\STATE \textbf{for }$i\in[1,\dots,I]$\textbf{ do }
			\STATE \hspace{0.5cm} $\mathrm{RandSelect}([1,\dots,N],b)\rightarrow S_i$
			\STATE \hspace{0.5cm} \textbf{for }$n\in S_i$\textbf{ do }
				\STATE \hspace{1cm} $\mathrm{Train}(M_{i-1},D_n)=M_i^{(n)}$
			\STATE \hspace{0.5cm} $M_i=\frac1b\sum_{n\in S_i}M_i^{(n)}$
			\STATE \hspace{0.5cm} $\omega_i=\mathrm{Acc}(M_i,D_{N+1})-\mathrm{Acc}(M_{i-1},D_{N+1})$
			\STATE \hspace{0.5cm} \textbf{if }$i>1\text{ and }\omega_i>\omega_{i-1}$\textbf{ then }
				\STATE \hspace{1cm} \textbf{for }$n\in S_i$\textbf{ do }$\varphi_n\leftarrow \varphi_n+1$
				\STATE \hspace{1cm} \textbf{for }$n\in S_{i-1}$\textbf{ do }$\varphi_n\leftarrow \varphi_n-1$
			\STATE \hspace{0.5cm} \textbf{if }$\omega_i<0$\textbf{ then }
				\STATE \hspace{1cm} \textbf{for }$n\in S_i$\textbf{ do }$\varphi_n\leftarrow \varphi_n-1$
	\end{algorithmic}
\end{algorithm}

We present the pseudo-code of the whole process in Algorithm \ref{alg:fedscore}. We split the dataset randomly into $N+1$ parts (line 1), representing the $N$ datasets of the participants and the test set $D_{N+1}$, to determine the quality of the aggregated updates. As highlighted earlier, the splitting is done in a way that the resulting sub-datasets are IID; otherwise, the splitting itself would introduce some quality difference between the participants. 

Concerning $D_{N+1}$, having access to a dataset is standard practice both in the field of privacy attacks and contribution score computation, and our work is at the intersection of these. Shadow datasets are a widespread technique to mimic the training dataset, and having access to an evaluation oracle (via an IID test set) is a fundamental assumption for contribution score computation methods. Although we foresee multiple options for how $D_{N+1}$ could be obtained, this is orthogonal to our main contribution; we leave it as a future work. 

Next, we artificially create the baseline dataset qualities using Eq. \ref{eq:scramble} (line 3): each participant's labels are randomized with a different ratio. This is followed by FL (lines 5-9). Round-wise improvements are captured by $\omega$ (declared in line 11 using the accuracy difference of the current and previous models). Quality scores ($\varphi_1,\dots,\varphi_N$) are updated in the $i$th round with $\pm1$ each time one of the three scoring rules is invoked (line 12, 13, and 15 for \emph{The Good}, \emph{The Bad}, and \emph{The Ugly}, respectively).\ 

As a related side issue, we briefly study how this label-flipping strategy affects the final achieved accuracy of the model in Section \ref{sec:app} (i.e., see Figure \ref{fig:weight_small}).
\section{Experimental Results}
\label{sec:qimeasure}

In this section, we detail our experimental results and elaborate on possible rule improvements and plausible mitigation strategies. 

\begin{figure*}[t!]
	\centering
	\begin{subfigure}{\textwidth}
		\includegraphics[width=4.2cm]{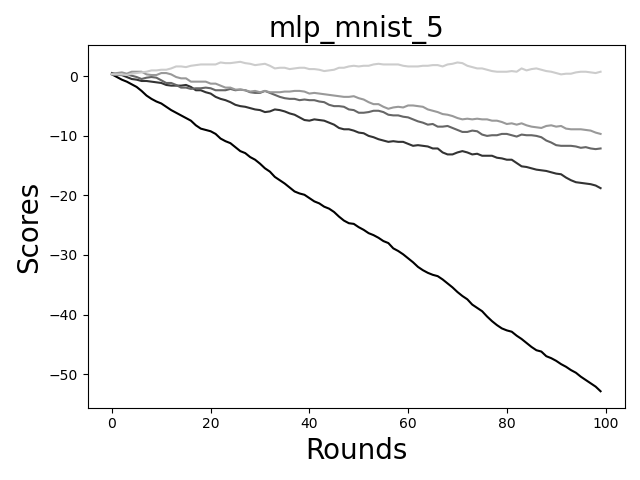}
		\includegraphics[width=4.2cm]{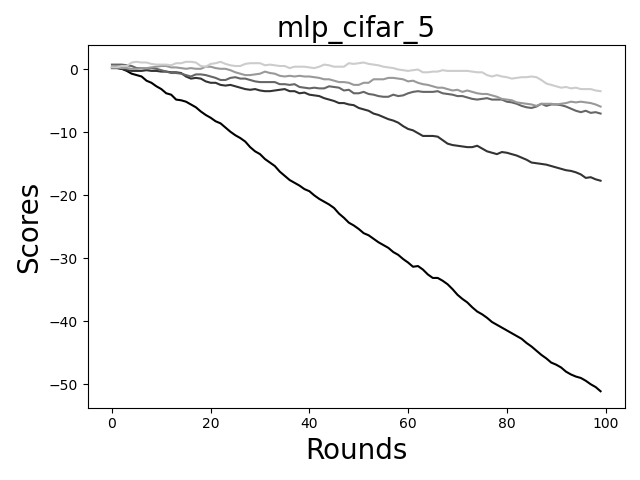}
		\includegraphics[width=4.2cm]{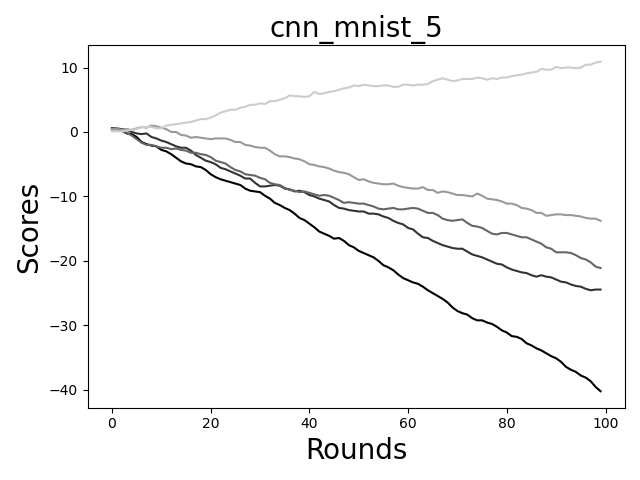}
		\includegraphics[width=4.2cm]{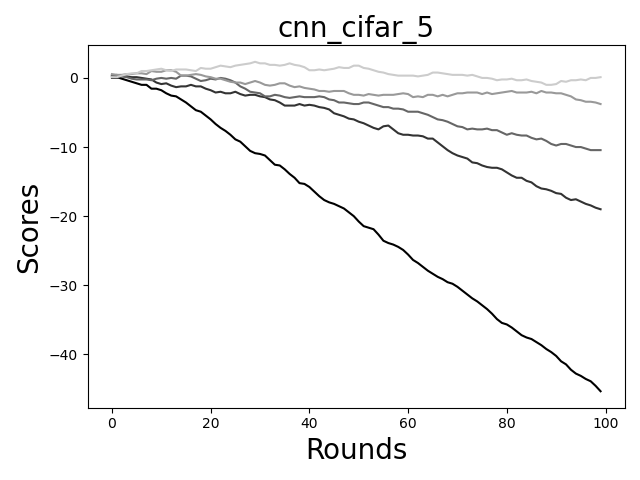}
		\vspace{0.5cm}
		\includegraphics[width=4.2cm]{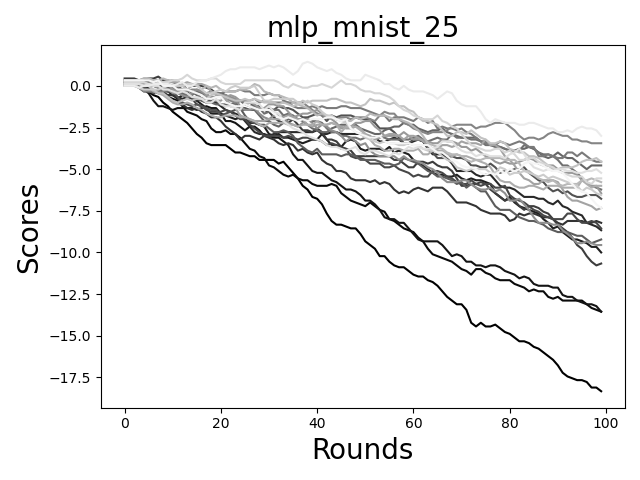}
		\includegraphics[width=4.2cm]{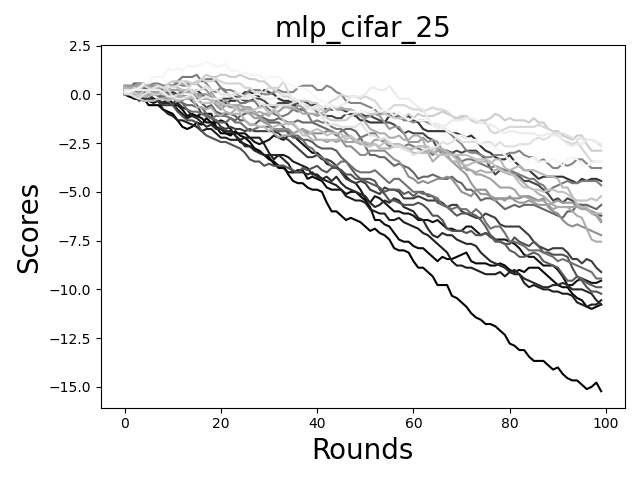}
		\includegraphics[width=4.2cm]{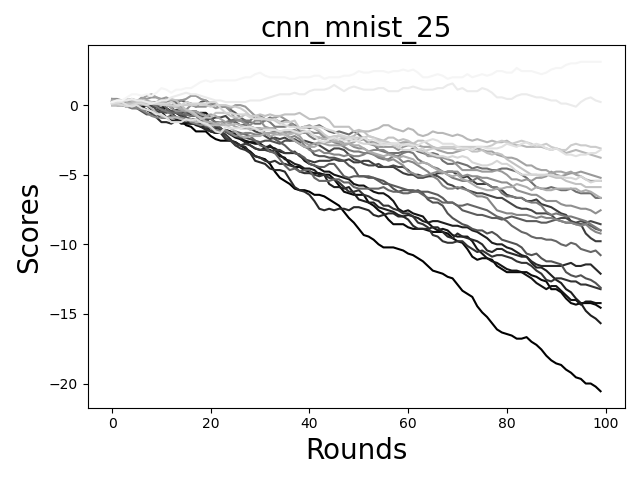}
		\includegraphics[width=4.2cm]{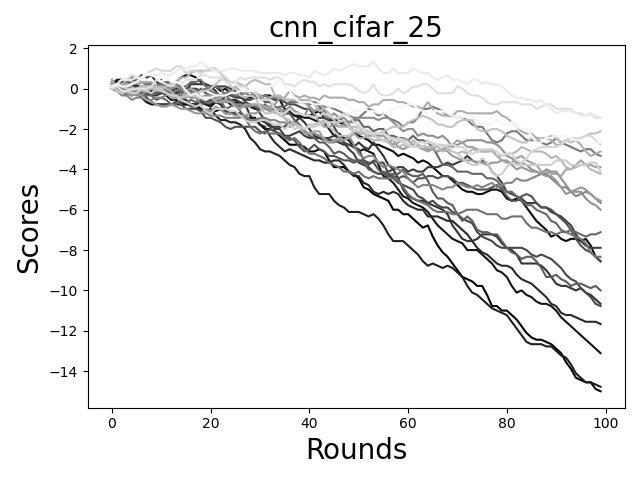}
		\vspace{0.5cm}
		\includegraphics[width=4.5cm]{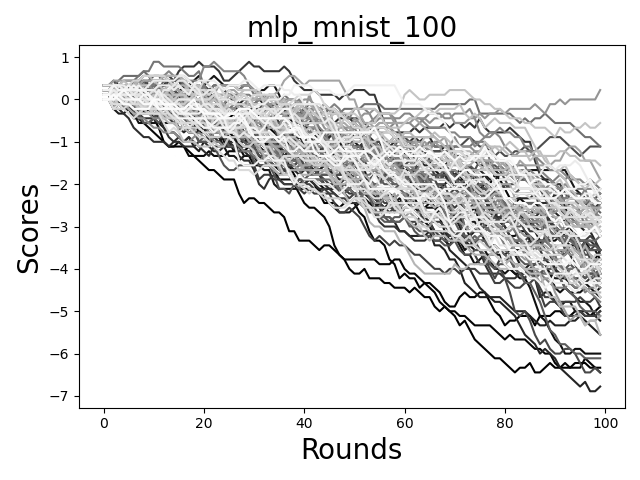}
		\includegraphics[width=4.5cm]{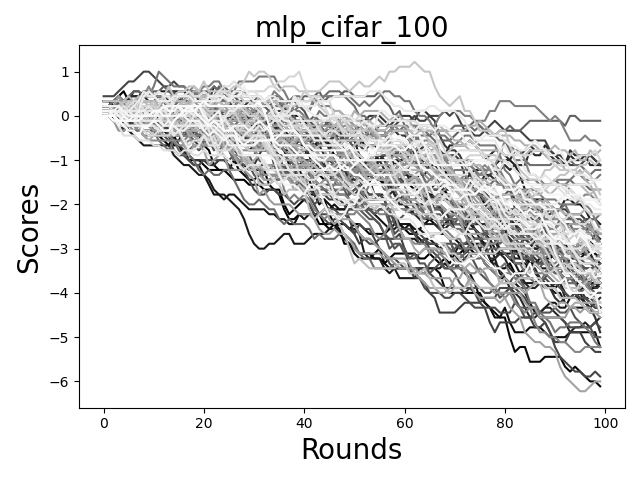}
		\includegraphics[width=4.5cm]{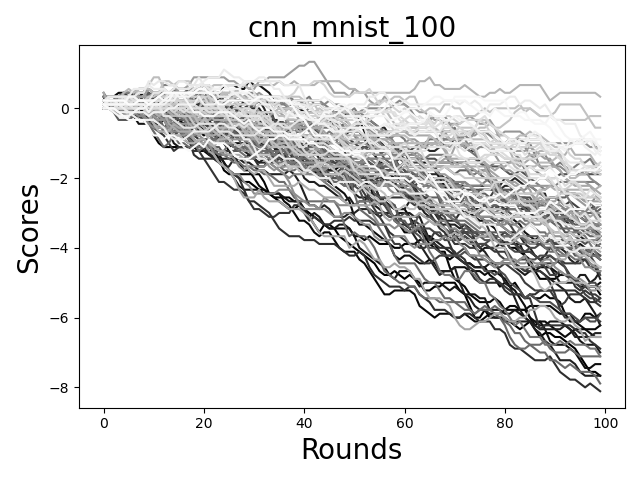}
		\includegraphics[width=4.5cm]{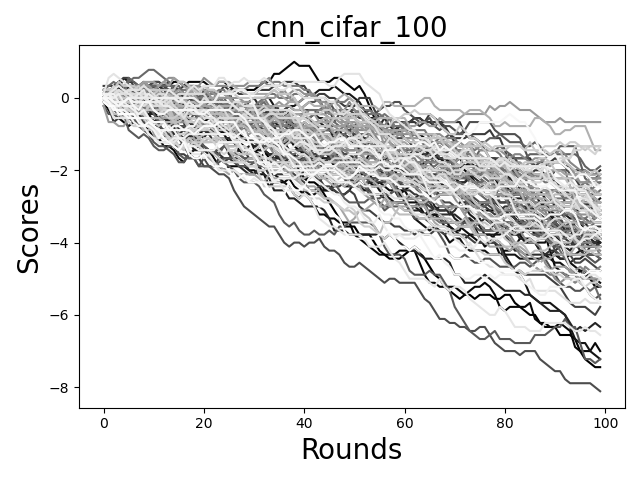}
		\vspace{0.5cm}
		\caption{The average round-wise change of the participants' scores. From bottom to top: 5, 25, and 100 participants. From left to right: MM, MC, CM, and CC. The lighter the better (the darker the worse) corresponding dataset quality.}
		\label{fig:QI}
	\end{subfigure}
	\begin{subfigure}{\textwidth}
		\includegraphics[width=4.2cm]{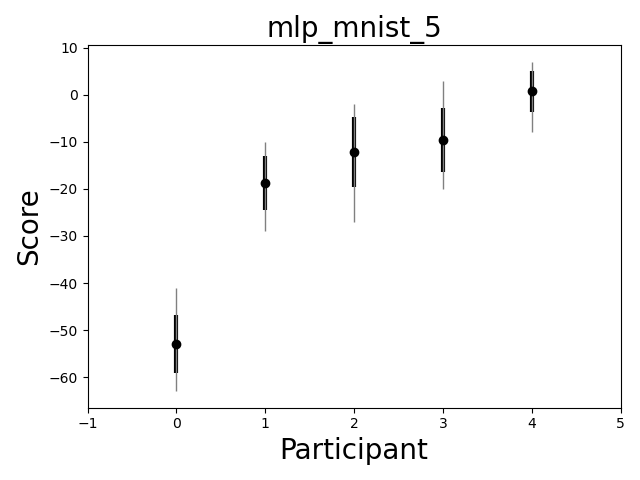}
		\includegraphics[width=4.2cm]{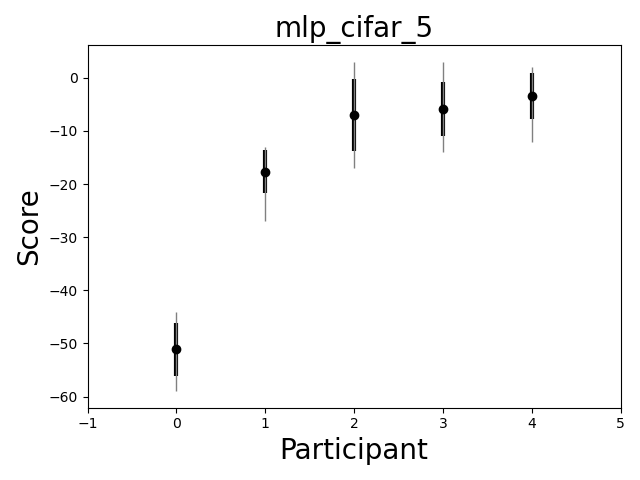}
		\includegraphics[width=4.2cm]{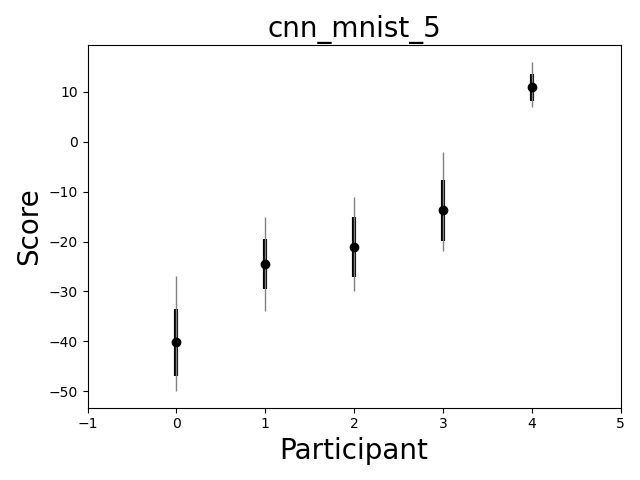}
		\includegraphics[width=4.2cm]{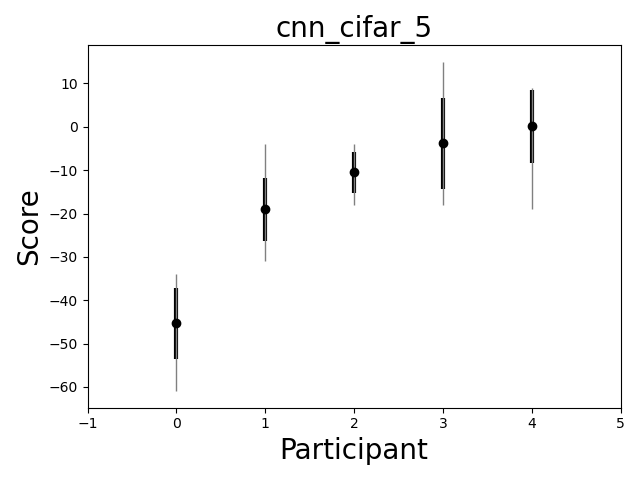}
		\vspace{0.5cm}
		\includegraphics[width=4.2cm]{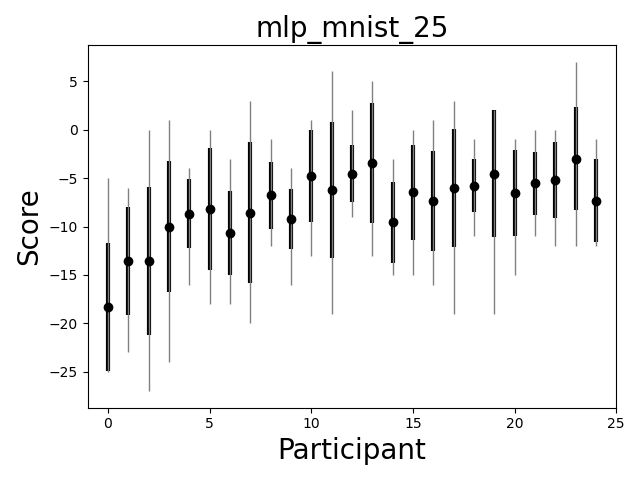}
		\includegraphics[width=4.2cm]{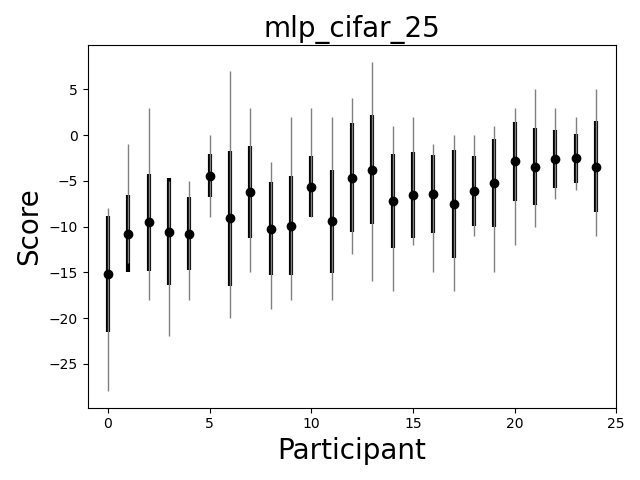}
		\includegraphics[width=4.2cm]{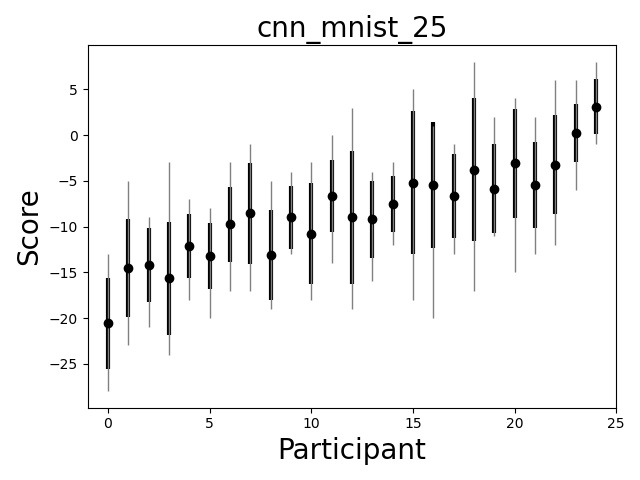}
		\includegraphics[width=4.2cm]{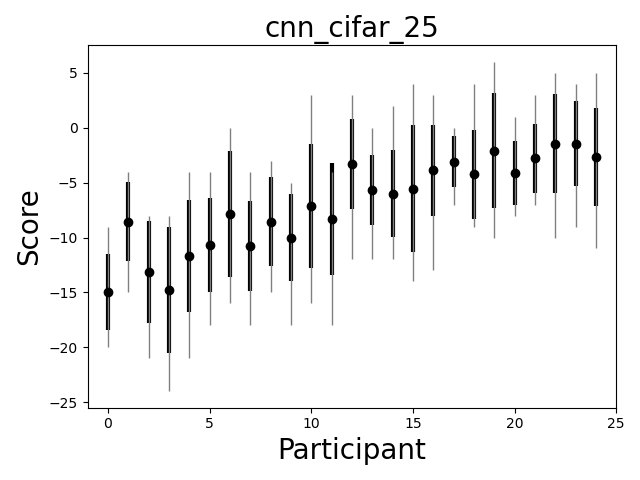}
		\vspace{0.5cm}
		\includegraphics[width=4.5cm]{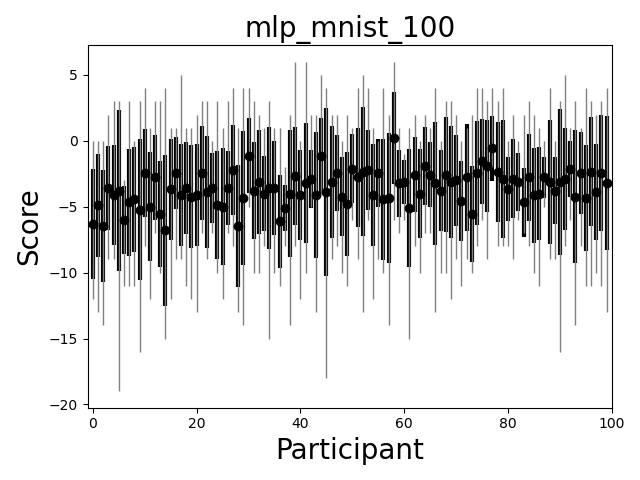}
		\includegraphics[width=4.5cm]{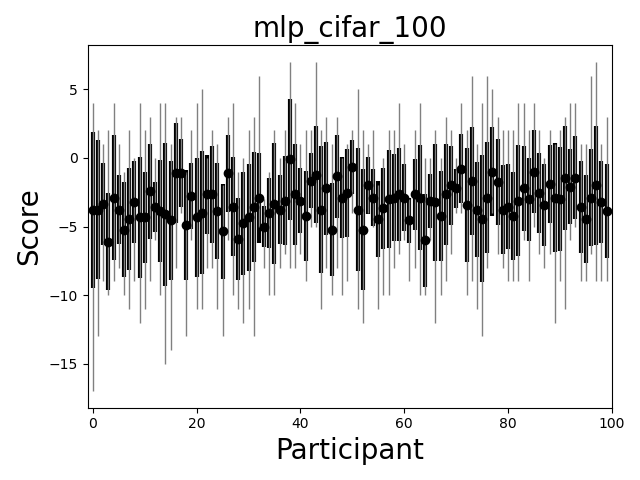}
		\includegraphics[width=4.5cm]{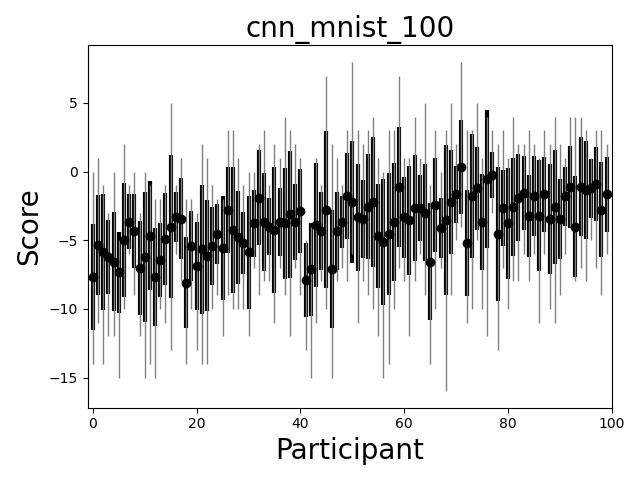}
		\includegraphics[width=4.5cm]{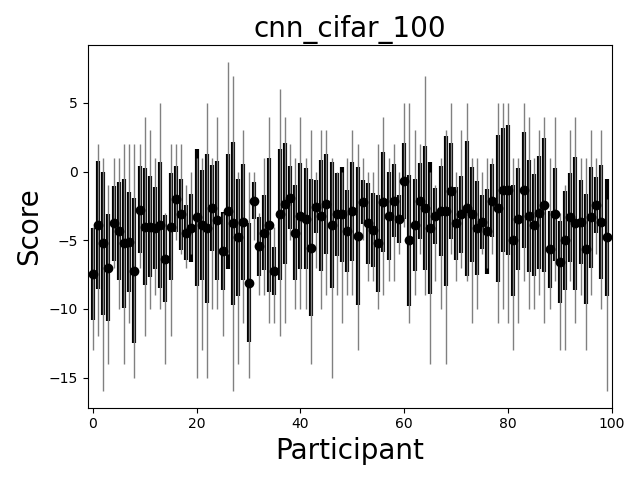}
		\vspace{0.5cm}
		\caption{Quality scores of the participants. From left to right: MM, MC, CM, and CC. From top to bottom: 5, 25, and 100 participants with IDs shown on $x$ axis where smaller numbers correspond to lower dataset quality. }
		\label{fig:boxplot}
	\end{subfigure}
\end{figure*}

The quality scores based on the three scoring rules are presented in Figure \ref{fig:QI} and Figure \ref{fig:boxplot}). In Figure  \ref{fig:QI} we visualize the round-wise evolution of scores for each participant where the corresponding grayness level depends on the participant ID. More precisely, the lighter shades correspond to participants with higher IDs (i.e., less noisy labels according to Eq. \ref{eq:scramble}), while the darker shades mark low ID participants (i.e., a higher ratio of random labels). It is visible that the more rounds have passed, the better our scoring rules correctly differentiate the participants.

In Figure  \ref{fig:boxplot} we show the mean (dot), the variance (black line), and the minimum and maximum values (gray line) of the inferred quality scores for each participant. One can see an increasing trend in the quality scores following the participant IDs. This is in line with the ground truth based on Eq. \ref{eq:scramble}. Note, that even for the participant with the perfect label quality (i.e., the highest ID or the lightest curve), the quality score is rather negative, and keeps decreasing with more rounds. This is an expected characteristic of the scoring rules: there is only one rule to increase the score (\emph{The Good}), while two to decrease it (\emph{The Bad} and \emph{The Ugly}). Applied jointly, these three heuristic scoring rules approximate the ground truth label quality ordering remarkably well \emph{exclusively from the aggregates}.

Finally, we utilize the Spearman coefficient $r_s$ introduced in Eq. \ref{eq:qi} to measure the accuracy of the inferred qualities; the 12 studied scenarios are presented in Figure  \ref{fig:qiter}. Note, that $r_s\in[-1,1]$, and any positive value indicates correlation. Thus, the value of the baseline (i.e., randomly guessed ordering) is zero. Consequently, the three simple rules significantly improve on the baseline, as the coefficients for all scenarios are positive. Moreover, as suggested by \ref{fig:QI}, this value keeps increasing with more rounds, as shown in the Appendix (Figure  \ref{fig:QI_boxplot}). 

\subsection{Fine-tuning}

We consider four ways of improving the accuracy of QI. 

\begin{itemize}
    \item \emph{Rule combination}: we apply all possible combinations of scoring rules in order to remove redundancies and to find which setup obtains the highest accuracy.  
    \item \emph{Thresholding}: we consider using a threshold for the scoring rules, i.e., \emph{The Ugly} only applies when the improvement is below some value, while \emph{The Good}/ \emph{The Bad} applies if the improvement difference is above/below such a threshold, respectively. 
    \item \emph{Actual values}: we consider using improvement differences instead of $\pm1$ to account for a more precise differentiation. 
    \item \emph{Round skipping}: In the early rounds the model does improve almost independently of the dataset qualities, therefore, we consider discarding the information from the first few rounds to decrease noise. 
\end{itemize}
	
\begin{figure}[t]
	\centering
	\begin{subfigure}{0.45\textwidth}
		\includegraphics[width=6cm]{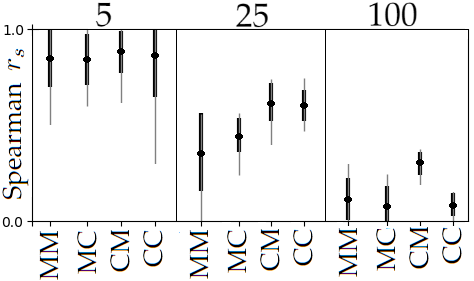}
		\caption{Spearman coefficient for the 12 scenarios.}
		\label{fig:qiter}
	\end{subfigure}
	\hfill
	\begin{subfigure}{0.45\textwidth}
		\includegraphics[width=8cm]{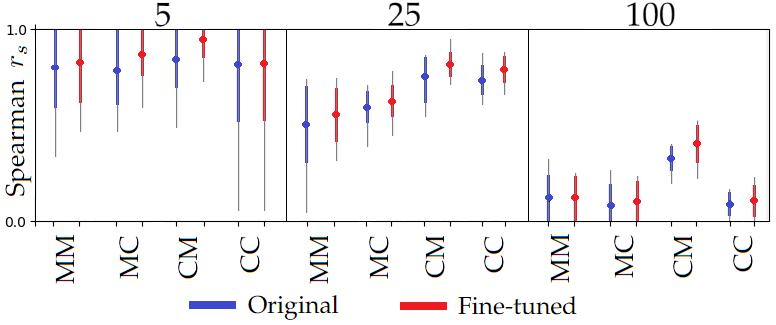}
		\caption{The original Spearman coefficient vs. the fine-tuned coefficient for the 12 scenarios. }
		\label{fig:opt}
	\end{subfigure}
\end{figure}

Although we performed an exhaustive grid search (e.g., $\{0, 2^0, \dots, 2^8\}/100$ for thresholding and $[0, 1, \dots, 10]$ for round skipping), the overall improvements obtained were minor. The corresponding results are presented in Figure \ref{fig:opt}. This implies that the original rules are quite efficient, and the heuristic thumbs-up/thumbs-down rules (e.g., using $\pm1$ to update the scores) could be interpreted as a normalizer across the different improvement levels of the rounds. Therefore, in the following applications, we use the original rules without any fine-tuning.  

\subsection{Mitigation}

Note that the demonstrated quality information leakage is not by design; this is a bug, rather than a feature in FL. The simplest and most straightforward way to mitigate this vulnerability is to use a protocol where every participant contributes in each round (incurring a sizable communication overhead). Another approach is to hide the participants' IDs (e.g., via mixnets \cite{chaum1981untraceable}), so no one knows which participant contributed in which round except for the participants themselves. Finally, the aggregation itself could be done in a differentially private manner as well, where a carefully calculated noise is added to the updates in each round. Client-level DP \cite{geyer2017differentially} would by default hide the dataset quality of the participants, although at the price of requiring large volumes of noise, and therefore, having low utility.

\section{Applications of QI}
\label{sec:app}

In this section, we envisage three scenarios where computing quality scores could be helpful: training accuracy stabilization, contribution score computation, and misbehavior detection. 

Even though QI is not a mechanism purposefully engineered into FL (with \emph{SA}), it does enable the above-mentioned beneficial applications. Note that while there are a handful of existing mechanisms for these tasks within FL, they do not work under \emph{SA}; hence, we do not compare our results quantitatively to the SotA methods. Our results are shown in Figure  \ref{fig:app}. 

\subsection{Enhancing the Training}

It is expected that both training speed and obtained accuracy could be improved by weighting the participants according to their data qualities. Hence, a potential use case for QI is to adopt the inferred scores as weights during training. For weighting, we used the multiplicative weight update approach \cite{arora2012multiplicative}, which multiplies the weights with a fixed rate $\kappa$, i.e., each time during training one of the three scoring rules is invoked in Algorithm \ref{alg:fedscore}, the weights (initialized as $[1,\dots,1]$) are updated in the $i$th round with $\times(1\pm\kappa)$ for the appropriate participants. 

\begin{figure*}[t]
	\centering
	\includegraphics[width=4.2cm]{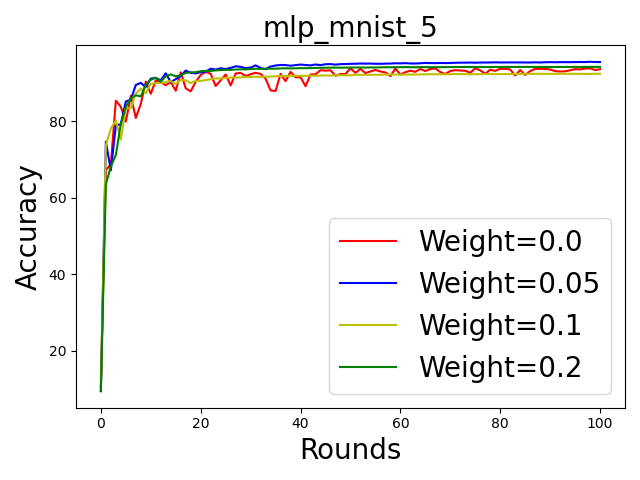}
	\includegraphics[width=4.2cm]{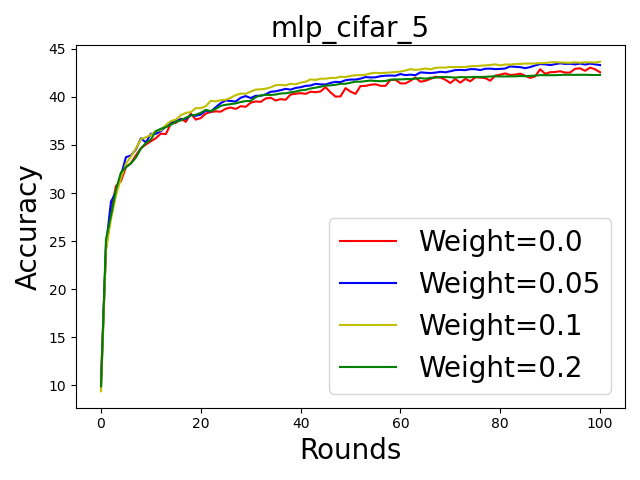}
	\includegraphics[width=4.2cm]{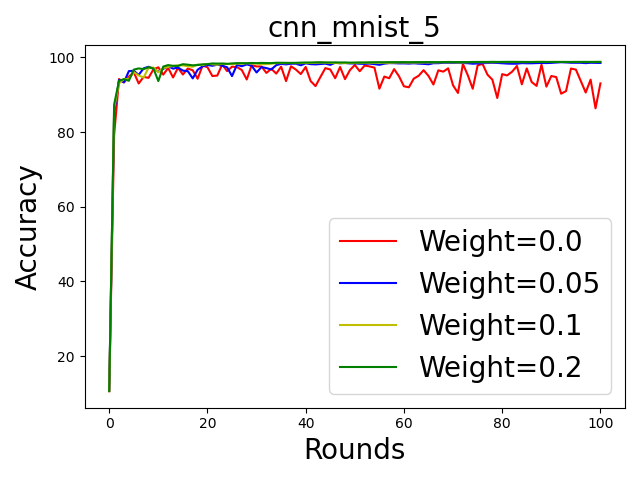}
	\includegraphics[width=4.2cm]{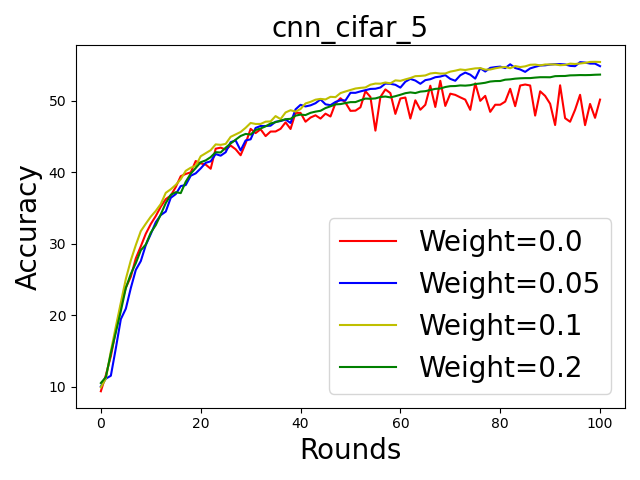}
	\vspace{0.5cm}
	\includegraphics[width=4.2cm]{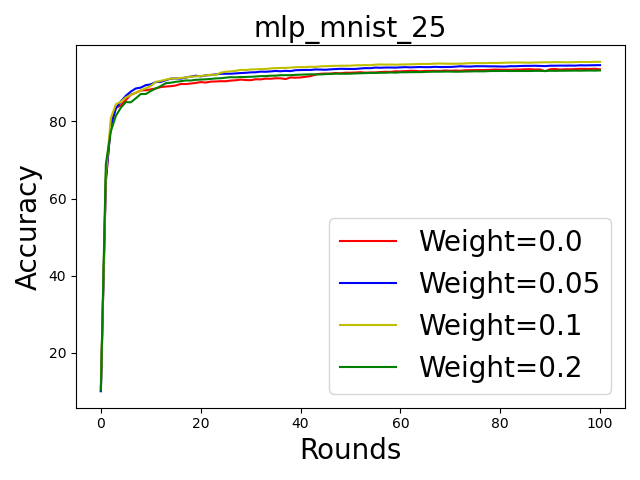}
	\includegraphics[width=4.2cm]{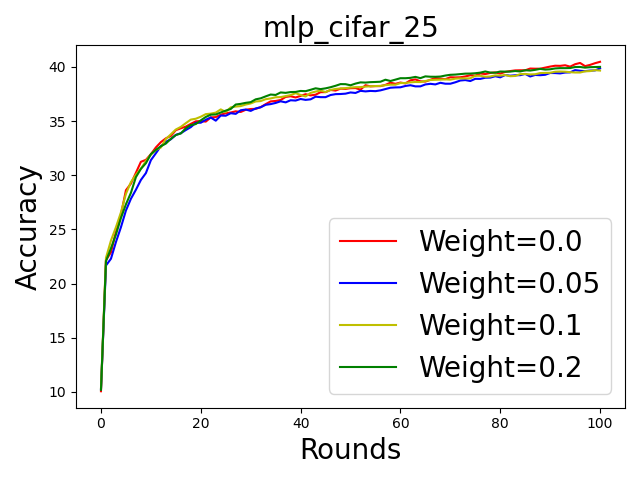}
	\includegraphics[width=4.2cm]{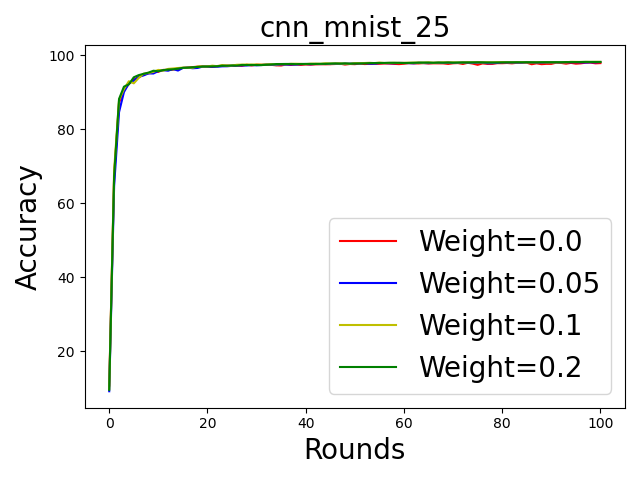}
	\includegraphics[width=4.2cm]{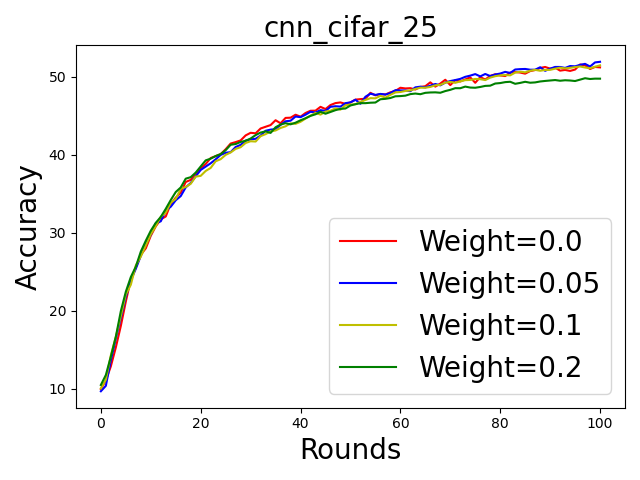}
	\vspace{0.5cm}
	\includegraphics[width=4.5cm]{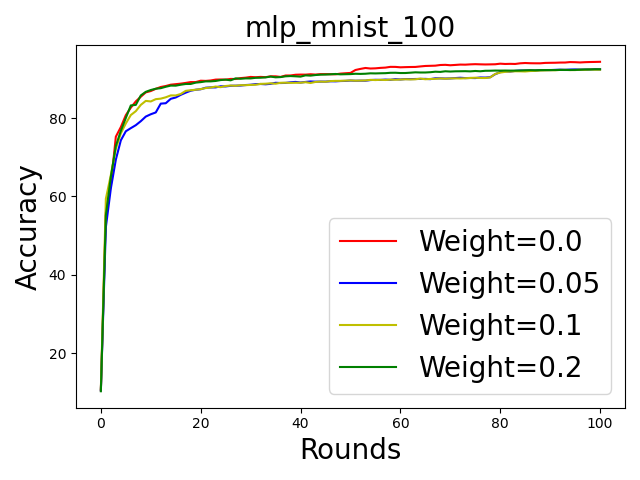}
	\includegraphics[width=4.5cm]{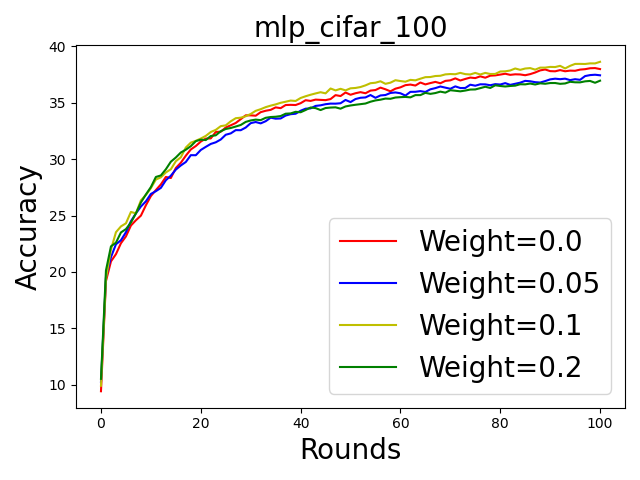}
	\includegraphics[width=4.5cm]{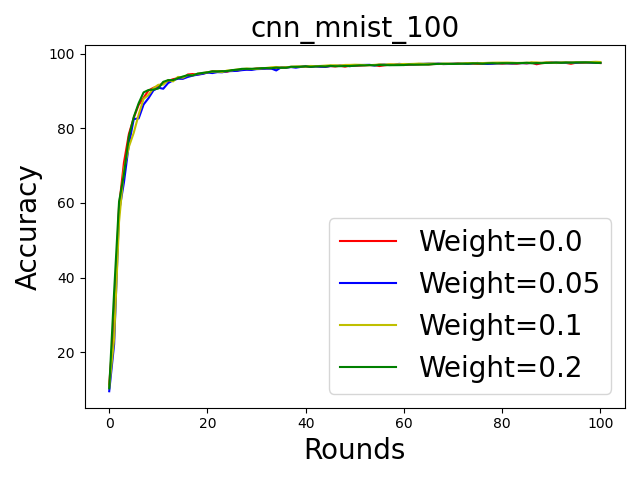}
	\includegraphics[width=4.5cm]{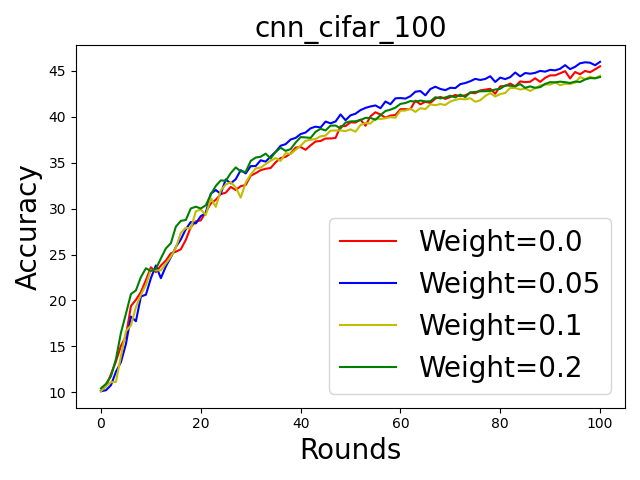}
	\vspace{0.5cm}
	\caption{The round-wise accuracy of the trained models with various weights. From top to bottom: MM, MC, CM, and CC, from left to right: 5, 25, and 100 participants.}
	\label{fig:weight}
\end{figure*}

Note that without access to individual gradients (owing to \emph{SA}), only the aggregates can be scaled by the server. Consequently, in each round, only the aggregate is scaled with the arithmetic mean of the selected participants' weights. For our experiments, we set $\kappa=\{0.00, 0.05, 0.10, 0.20\}$, where the first value corresponds to the baseline without participant weighting. We highlight some of our results in Figure \ref{fig:weight}. It is conclusive that using weights based on our scoring rules enhances the training as the training curves are smoother and the final accuracies are higher.

\subsection{Contribution Score Computation}

The second use case we envisioned for QI is contribution score computation. The holy grail of this sub-discipline is the Shapley value \cite{shapley1953value}, which is exponentially hard to compute, as besides the individual information, it requires information about all potential coalitions of participants. Thus, many approximation methods exist (e.g., \cite{ghorbani2019data,wang2020principled}. Yet, all methods assume explicit access to the individual datasets or the corresponding gradients, which is not possible with \emph{SA}. Consequently, there exists no contribution scoring mechanism which could be considered a relevant baseline for QI. 

According to \cite{huang2021shapley}, payment distribution based on the Shapley value is optimal for our IID setting. Moreover, the federated leave-one-out method (LO) method approximates the Federated Shapley value well, in this case \cite{wang2020principled}. Although LO does need individual information (hence, not applicable with \emph{SA}), we compare our method to it, as it only utilizes each individual gradient once (to obtain the grand coalition minus that participant).

\begin{figure}[t]
	\centering
	\begin{subfigure}{0.45\textwidth}
		\includegraphics[width=6cm]{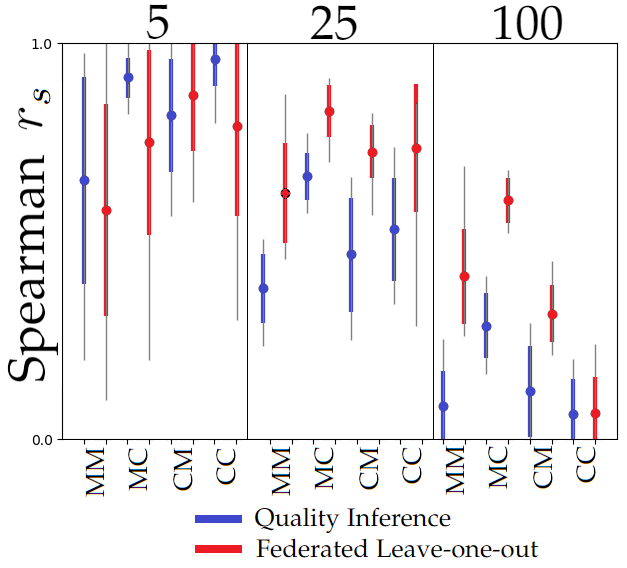}
		\caption{Spearman coefficient of QI and the leave-one-out method for the 12 scenarios.}
		\label{fig:loo}
	\end{subfigure}
	\hfill
	\begin{subfigure}{0.45\textwidth}
		\includegraphics[width=8cm]{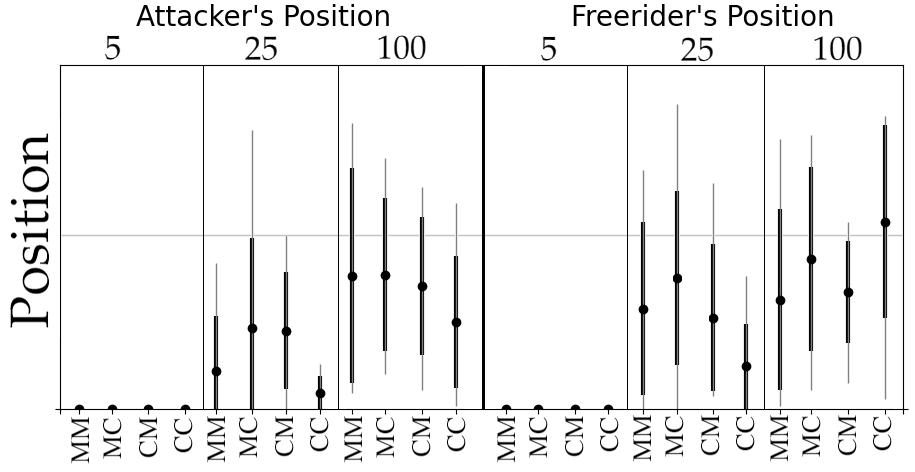}
		\caption{Position of a single attacker (left) and free-rider (right) for the twelve use-cases after hundred training rounds. The higher/lower results correspond to higher/lower inferred quality-wise ranks.}
		\label{fig:QI_attack}
	\end{subfigure}
\end{figure}

The Spearman coefficients of the ordering based on QI and LO are presented in Figure \ref{fig:loo}. As expected, LO is superior to QI, as it operates on individual information, which is by-design avoided by QI. What is somewhat surprising is that LO (benefiting from individual gradients) also struggles with reconstructing the quality-wise ordering perfectly. This suggests that separating participants with different label qualities is indeed a challenging task; given the restricted information setting, QI performs reasonably well. 

\subsection{Misbehavior Detection}

\begin{table}[!b]
    \centering
   	\resizebox{0.45\textwidth}{!}{%
    \begin{tabular}{|cc|cccc|}
        \hline
         Setup & Attacker & MM & MC & CM & CC \\
         \hline
         5/2/1 & p-value &2.0e-20 & 3.5e-37 & 8.3e-58 & 5.4e-84\\
         T-Test & Stat &16.4 & 17.7 & 21.9 & 27.0\\
         \hline
         25/5/2 & p-value  &6.8e-06 & 1.9e-10 & 1.1e-14 & 6.0e-27\\
         $\chi^2$-Test & Stat &38.3 & 62.0 & 83.3 & 146.0 \\
         \hline
         100/10/5 & p-value &8.6e-03& 1.2e-03& 1.7e-07& 1.1e-08\\
         KS-Test & Stat &0.25 &0.17& 0.19& 0.18 \\
        \hline    
    \end{tabular}}
	\resizebox{0.45\textwidth}{!}{%
    \begin{tabular}{|cc|cccc|}
        \hline
        Setup & Free-Rider& MM & MC & CM & CC \\
        \hline
        5/2/1 & p-value & 3.7e-21 & 1.6e-42 & 4.3e-69 & 5.1e-98\\
        T-Test & Stat & 12.3 & 18.1 & 24.0 & 29.5\\
        \hline
        25/5/2 & p-value & 7.7e-03 & 7.5e-12 & 7.0e-17 & 6.1e-39\\
        $\chi^2$-Test & Stat & 20.8 & 69.1 & 96.8 & 203.5\\
        \hline
        100/10/5 & p-value&9.0e-02& 2.0e-05 &3.6e-07 &4.8e-14\\
        KS-Test & Stat & 0.13& 0.18& 0.17& 0.21 \\
        \hline
    \end{tabular}}
    \caption{Statistics and p-values of the selected scenarios: X/Y/Z mean number of participants, number of round-wise selected participants, and number of cheaters, respectively. It is clear, that the hypothesis ``the scores of honest and cheating participants are similar'' (e.g., having the same mean for the Student T-Test, having the same frequencies for the $\chi^2$-Test, and coming from the same distribution for the Kolmogorov-Smirnov Test) is rejected with high confidence.}
    \label{tab:score}
\end{table}

Another potential application of QI is misbehavior detection. It is a notoriously hard task even without \emph{SA} \cite{fung2018mitigating}. At the time of writing, we are not aware of any work tackling this problem in the \emph{SA} setting. 

Here we consider both malicious attackers and free-riders. Their goal is either to decrease the accuracy of the aggregated model or to benefit from the aggregated model without contributing, respectively. We do not scramble the labels of honest participants, and simulate attackers by computing the additive inverse of the correct gradients, while we use zero as the gradient for free-riders. These are naive but stealthy strategies owing to \emph{SA}. With this use case, our goal is not to propose a defense against SotA attackers but rather to demonstrate the usability of QI besides label quality inference. Note that QI also shows promise for being applicable to determine other quality disparities among participants. 

We studied the score of the honest and malicious participants; the average values for the selected scenarios are presented in Table \ref{tab:score}; the rest can be found in the Appendix (Table \ref{tab:cheating}). We also run various statistical tests to determine whether there is any difference between the honest and malicious participant's scores. Table \ref{tab:score} contains highlighted results, while the rest is presented in the Appendix (Table \ref{tab:t-test}, \ref{tab:w-test}, \ref{tab:wm-test}, \ref{tab:cs-test}, and \ref{tab:ks-test}). The tests concluded unanimously that the two score distributions are different, thus, QI is capable of correctly flagging dishonest participants. Besides the score differences, we also studied the inferred position of a single cheater, which is always in the bottom half (see Figure \ref{fig:QI_attack}).
\section{Related Work}
\label{sec:rw}

In this section, we briefly present related research efforts, including but not limited to data quality scoring mechanisms and well-known privacy attacks against machine learning. The theoretical analysis of QI does relate to \cite{dinur2003revealing} as attempting to reconstruct the dataset quality order is similar to reconstructing the entire dataset based on query outputs. 

\subsection{Participant Scoring}

Simple but effective scoring rules are prevalent in complex ICT-based systems, especially characterizing quality. For instance, binary or counting signals can be utilized to i) steer peer-to-peer systems measuring the trustworthiness of peers \cite{kamvar2003incentives}, ii) assess and promote content in social media \cite{van2011human}, iii) ensure the proper natural selection of products in online marketplaces \cite{lim2010detecting}, and iv) select trustworthy clients via simple credit scoring mechanisms \cite{thomas2017credit}.

There exist free-rider detection mechanisms for collaborative learning \cite{lin2019free,fraboni2021free}. In contrast, \cite{liu2021real} proposes an online evaluation method that defines each participant's impact based on the current and the previous rounds. Although their goal is similar to ours, we consider \emph{SA} being utilized, while neither of the above mechanisms is applicable in such a case. A disaggregation technique is presented in \cite{so2021securing}, which reconstructs the participation matrix by simulating the same round several times with different participants. Instead, we assume such participation information to be available and emulate the training rounds by properly updating the model. 

Accuracy boosting by participant weighting is considered in \cite{chen2020focus} where the weights are determined by the underlying data quality calculated via the cross-entropy of the local model predictions. These experiments consider only five participants and two quality classes (fully correct or incorrect); we study fine-grained quality levels with larger sets of participants. A similar method was utilized in an \emph{SA} setting in \cite{guo2020secure} using homomorphic encryption. In contrast, our method does not require any cryptographic primitive and can be utilized on top of any federated learning protocol.  

We naively assume that data quality is directly related to the noise present in the labels. Naturally, this is a simplification: there is an entire computer science discipline devoted to data quality \cite{batini2016data}. 

Authors of \cite{huang2020exploratory} listed several incentive mechanisms for contribution computation in FL (which can be interpreted as data quality). A pertinent notion is the Shapley value \cite{shapley1953value}, which was designed to allocate goods to players proportionally to their contributions. A high-level summary of the role of the Shapley value within ML is presented in \cite{rozemberczki2022shapley}. The main drawback of the Shapley value is its exponential computational requirement, which makes it unfeasible in most scenarios. Several approximation methods were proposed in the literature using sampling \cite{castro2009polynomial}, gradients \cite{ghorbani2019data,nagalapatti2021game,ghorbani2020distributional,kwon2021beta,liu2021gtg,yang2022wt} and influence functions \cite{koh2017understanding,xue2020toward,xu2020data}. Although some are promising (e.g., the conceptual idea in \cite{pejomeasuring}), all previous methods assume explicit access to either the datasets or the corresponding gradients. Consequently, these methods are not applicable when \emph{SA} is enabled during FL. QI can be considered as the first step towards a contribution score when \emph{no information on individual datasets is available}. 

\subsection{Privacy Attacks}

There are several indirect threats against FL models.
These could be categorized into model inference \cite{fredrikson2015model}, membership inference \cite{shokri2017membership}, parameter inference \cite{tramer2016stealing}, and property inference \cite{melis2019exploiting}. QI could be considered as an instance of the last. 
Source inference \cite{hu2021source} is also such an attack, which could tie the extracted information to specific participants of FL. However, it does not work with \emph{SA}. 
Another property inference attack is the quantity composition attack \cite{wang_eavesdrop_2019}, which aims at inferring the proportion of training labels among the participants in FL. This attack is successful even under \emph{SA} protocols or DP. In contrast to our work, the paper focuses on inferring the distributions of the non-IID datasets while we aim to recover the relative quality information on IID datasets. 
Finally, \cite{wang2019beyond} also attempts to explore user-level privacy leakage within FL. Similarly to our work, the attack defines client-dependent properties, which then can be used to distinguish the clients from one another. The authors assume an active malicious server utilizing a computationally heavy GAN for the attack, which is the exact opposite of our honest-but-curious setup with limited computational power.

\subsection{Privacy Defenses}

QI can be considered as a property inference attack; hence, naturally, it can be ``mitigated'' via client-level DP \cite{geyer2017differentially}. Moreover, as we simulate different dataset qualities with the amount of added noise, we want to prevent the leakage of the added noise volume. Consequently, this problem relates to private privacy parameter selection, as label perturbation \cite{papernot2016semi} (which we use to mimic different dataset quality levels) is one technique for achieving DP \cite{desfontaines2020sok}. Although some works set the privacy parameter using economic incentives \cite{pejo2019together}, we are not aware of any research aiming to define the privacy parameter itself also privately. 
\section{Conclusion}
\label{sec:con}

Federated learning is the most popular collaborative learning framework, wherein each round only a subset of participants updates a joint machine learning model. Fortified with secure aggregation, only aggregated information is learned both by the participants and the server. Yet, in this paper, we devised a simple set of quality scoring rules that successfully recover the relative ordering of the participant's dataset qualities (measured by perturbed label ratio). Besides a small representative dataset to evaluate the improvement of the model after each aggregation, our method neither requires any computational power nor background information. 

Through a series of image recognition experiments, we showed that it is possible to restore the relative ordering based on label quality with reasonably high accuracy. Our experiments also revealed a connection between the accuracy of the quality inference and both the complexity of the task and the used architecture. Moreover, we performed an ablation study suggesting that the original rules are near optimal. Lastly, we demonstrated how quality inference could i) boost training efficiency by weighting the participants, ii) yield an operational contribution metric, and iii) detect misbehaving participants based on their quality scores.

\subsection*{Limitations and Future Work}

This paper has barely scratched the surface of quality inference in federated learning based only on aggregated updates. We foresee multiple avenues towards improving and extending this work, e.g., using machine learning techniques to replace our naive rules by relaxing the attacker constraints concerning computational power and background knowledge. In the early rounds, selecting the participants in a non-random manner similar to \cite{liu2021gtg} could also be beneficial. 

For clarity, we have restricted our experiments to visual recognition tasks with noisy labels as the measure of data quality. Although we expect our results to generalize well to other domains, we leave further experiments as future work. Finally, the personal data protection implications of the information leakage caused by quality inference are also of interest: should such quality information be considered private, and, consequently, should it fall under data protection regulations such as the GDPR? This issue has significant practical relevance to federated learning platforms already in operation.

\subsection*{Acknowledgement}

The authors are grateful to Andr{\'a}s T{\'o}tth for his work in the experiments on contribution score computation.
This work was funded in part by the European Union (Grant Agreement Nr. 10109571, SECURED Project); Project no. 138903, implemented with the support provided by the Ministry of Innovation and Technology from the NRDI Fund, financed under the FK\_21 funding scheme; and Project no. TKP2021-NVA-02, implemented with the support provided by the Ministry of Culture and Innovation of Hungary from the National Research, Development and Innovation Fund, financed under the TKP2021-NVA funding scheme.

\bibliographystyle{plain}
\bibliography{mybibfile}

\appendix

\begin{table*}[h]
    \centering
    \parbox{.48\linewidth}{
   	\resizebox{0.48\textwidth}{!}{%
    \begin{tabular}{|cc|cccc|cccc|}
        \hline
         \multirow{2}{*}{Scenario} & \multirow{2}{*}{Type} & \multicolumn{4}{|c|}{Attacker} & \multicolumn{4}{c|}{Free-Rider} \\
         && MM & MC & CM & CC & MM & MC & CM & CC \\
         \hline
         \multirow{2}{*}{5/2/1} & Honest &-2.97	&-2.97	&-4.06	&-4.78	&-3.58	&-3.44	&-4.19	&-4.53\\
         & Cheater &-28.78	&-30.33	&-31.33	&-27.78	&-13.00	&-16.00	&-20.33	&-25.00\\
         \hline
         \multirow{2}{*}{25/5/1} & Honest &-5.38	&-6.07	&-6.74	&-7.12	&-5.33	&-6.49	&-6.78	&-7.76\\
         & Cheater &-13.67	&-10.56	&-11.56	&-26.33	&-8.11	&-6.44	&-10.11	&-14.33\\
         \hline
         \multirow{2}{*}{25/5/2} & Honest &-1.41	&-1.95	&-2.24	&-2.32	&-1.52	&-1.86	&-2.62	&-2.82\\
         & Cheater &-5.44	&-7.61	&-5.06	&-11.33	&-1.17	&-4.39	&-5.15	&-7.28\\
         \hline
         \multirow{2}{*}{100/10/1} & Honest &-8.36	&-9.13	&-9.38	&-10.18	&-8.29	&-8.81	&-9.55	&-10.24\\
         & Cheater &-9.78	&-14.00	&-15.11	&-14.89	&-12.56	&-6.44	&-12.00	&-9.44\\
         \hline
         \multirow{2}{*}{100/10/2} & Honest &-7.89	&-9.00	&-9.72	&-10.12	&-8.24	&-9.24	&-9.37	&-10.32\\
         & Cheater &-10.78	&-11.39	&-16.00	&-15.22	&-10.17	&-13.28	&-10.44	&-13.78\\
         \hline
         \multirow{2}{*}{100/10/3} & Honest &-7.96	&-8.98	&-9.50	&-10.08	&-8.77	&-9.25	&-9.51	&-9.96\\
         & Cheater &-12.52	&-11.63	&-12.85	&-18.19	&-8.56	&-9.48	&-12.00	&-13.67\\
         \hline
         \multirow{2}{*}{100/10/4} & Honest &-8.07	&-8.93	&-9.52	&-9.63	&-8.33	&-9.06	&-9.57	&-10.08\\
         & Cheater &-11.31	&-12.67	&-13.47	&-14.33	&-9.86	&-10.53	&-12.17	&-12.08\\
         \hline
         \multirow{2}{*}{100/10/5} & Honest &-8.17	&-8.67	&-9.57	&-10.01	&-8.38	&-8.88	&-9.57	&-9.68\\
         & Cheater &-11.69	&-12.67	&-13.78	&-16.47	&-8.47	&-12.13	&-10.76	&-14.96\\
        \hline
    \end{tabular}}
    \caption{The average QI scores across several scenarios for the honest and cheating participants. X/Y/Z mean number of participants, number of round-wise selected participants, and number of cheaters, respectively. Attacker and free-rider using the additive inverse of the correct gradient and zero as gradient, respectively. MM, MC, CM, and CC stand for MNIST/MLP, MNIST/CNN, CIFAR/MLP, and CIFAR/CNN, respectively. }
    \label{tab:cheating}}
	\hfill
	\parbox{.48\linewidth}{
	\resizebox{0.48\textwidth}{!}{%
    \begin{tabular}{|cc|cccc|cccc|}
        \hline
         \multirow{2}{*}{Scenario} & \multirow{2}{*}{Type} & \multicolumn{4}{|c|}{Attacker} & \multicolumn{4}{c|}{Free-Rider} \\
         && MM & MC & CM & CC & MM & MC & CM & CC \\
         \hline
         \multirow{2}{*}{5/2/1} & p-value &2.0e-20 & 3.5e-37 & 8.3e-58 & 5.4e-84 & 3.7e-21 & 1.6e-42 & 4.3e-69 & 5.1e-98\\
         & Stat &16.4 & 17.7 & 21.9 & 27.0 & 12.3 & 18.1 & 24.0 & 29.5\\
         \hline
         \multirow{2}{*}{25/5/1} & p-value &9.2e-07 & 4.7e-07 & 4.4e-07 & 7.1e-18 & 6.1e-06 & 9.9e-06 & 7.6e-08 & 2.0e-20\\
         & Stat &4.9 & 4.9 & 4.9 & 8.6 & 4.4 & 4.3 & 5.3 & 9.3\\
         \hline
         \multirow{2}{*}{25/5/2} & p-value &2.9e-08 & 1.8e-11 & 2.0e-15 & 6.6e-30 & 4.8e-04 & 6.3e-13 & 1.2e-16 & 5.0e-35\\
         & Stat &5.6 & 6.7 & 8.0 & 11.5 & 3.3 & 7.2 & 8.3 & 12.6\\
         \hline
         \multirow{2}{*}{100/10/1} & p-value &2.5e-01& 1.9e-03& 1.7e-03 &1.0e-04& 2.7e-02 &2.7e-02 &8.9e-04& 4.5e-04\\
         & Stat &0.69& 2.89& 2.93& 3.71 &1.93 &1.94 &3.13 &3.32\\
         \hline
         \multirow{2}{*}{100/10/2} & p-value &2.3e-02& 2.4e-03 &1.6e-07 &2.6e-09 &9.8e-03 &8.7e-05 &3.0e-07& 2.6e-10\\
         & Stat &1.99& 2.83& 5.20& 5.85& 2.33& 3.76 &5.00 &6.22\\
         \hline
         \multirow{2}{*}{100/10/3} & p-value &6.6e-05& 4.6e-04 &7.3e-05& 8.2e-11& 5.7e-03 &1.7e-03& 9.4e-06 &1.1e-12\\
         & Stat &3.84 &3.32 &3.80 &6.40& 2.53& 2.92& 4.28 &7.03\\
         \hline
         \multirow{2}{*}{100/10/4} & p-value &1.3e-03 &3.3e-06& 5.3e-09 &1.1e-13& 1.0e-03& 2.0e-06 &3.3e-10& 3.7e-14\\
         & Stat &3.02& 4.52& 5.73& 7.35 &3.08& 4.62& 6.18 &7.50\\
         \hline
         \multirow{2}{*}{100/10/5} & p-value &1.5e-04 &3.9e-06 &2.1e-12& 1.2e-18 &4.3e-03& 1.3e-08 &5.1e-12&6.7e-24\\
         & Stat &3.64& 4.48 &6.95 &8.76& 2.63& 5.58 &6.82 &10.05\\
        \hline
    \end{tabular}}
    \caption{Statistics and p-values of the Student T-test with hypothesis: score of honest and cheating participants have the same mean. Results suggest rejecting this hypothesis for most scenarios with high confidence, hence, there is a significant difference in the obtained quality scores between the two groups. }
    \label{tab:t-test}}
\end{table*}

\begin{table*}[h]
    \centering
    \parbox{.48\linewidth}{
   	\resizebox{0.48\textwidth}{!}{%
    \begin{tabular}{|cc|cccc|cccc|}
        \hline
         \multirow{2}{*}{Scenario} & \multirow{2}{*}{Type} & \multicolumn{4}{|c|}{Attacker} & \multicolumn{4}{c|}{Free-Rider} \\
         && MM & MC & CM & CC & MM & MC & CM & CC \\
         \hline
         \multirow{2}{*}{5/2/1} & p-value &1.0e-09 & 3.1e-13 & 3.6e-21 & 3.1e-31 & 6.1e-08 & 4.1e-16 & 1.6e-25 & 2.4e-37\\
         & Stat &15.9 & 12.1 & 15.5 & 19.8 & 8.2 & 12.8 & 17.2 & 22.2\\
         \hline
         \multirow{2}{*}{25/5/1} & p-value &1.1e-03 & 6.8e-05 & 1.1e-05 & 2.6e-08 & 5.2e-04 & 2.2e-04 & 1.3e-06 & 6.3e-10\\
         & Stat &4.3 & 4.4 & 4.7 & 6.2 & 3.9 & 3.9 & 5.2 & 6.9 \\
         \hline
         \multirow{2}{*}{25/5/2} & p-value &2.1e-05 & 3.2e-07 & 5.6e-10 & 1.1e-15 & 4.1e-03 & 1.2e-08 & 4.3e-11 & 2.5e-19\\
         & Stat &5.2 & 5.6 & 6.7 & 8.9 & 2.8 & 6.2 & 7.1 & 10.2\\
         \hline
         \multirow{2}{*}{100/10/1} & p-value &2.2e-01& 3.6e-04 &1.3e-03& 2.7e-04 &2.1e-02 &1.4e-02 &6.8e-04& 6.8e-04\\
         & Stat &0.96& 3.82 &3.19 &3.65 &2.20 &2.30& 3.38& 3.33\\
         \hline
         \multirow{2}{*}{100/10/2} & p-value &2.2e-02 &9.5e-04 &3.0e-06& 8.0e-08& 6.3e-03& 4.6e-05 &4.3e-06& 3.9e-09\\
         & Stat &2.17& 3.26& 4.80& 5.54& 2.62& 4.14& 4.67 &6.13\\
         \hline
         \multirow{2}{*}{100/10/3} & p-value &5.3e-04& 4.2e-04& 9.0e-05& 2.2e-09& 6.5e-03 &1.3e-03& 7.0e-06& 2.1e-11\\
         & Stat &3.66 3.46& 3.85& 6.14& 2.56& 3.07 &4.47& 6.94 \\
         \hline
         \multirow{2}{*}{100/10/4} & p-value &8.9e-04 &2.6e-06& 1.1e-08 &6.7e-13 &6.5e-04 &1.2e-06& 4.9e-10 &1.1e-13\\
         & Stat &3.35& 4.78& 5.84& 7.44& 3.34 &4.89& 6.37 &7.67\\
         \hline
         \multirow{2}{*}{100/10/5} & p-value &2.1e-04 &1.1e-05 &1.6e-11 &1.4e-14& 5.2e-03& 4.7e-08 &2.3e-11 &1.2e-19\\
         & Stat &3.78 &4.39& 6.95 &7.95& 2.61& 5.55& 6.84& 9.48\\
        \hline
    \end{tabular}}
    \caption{Statistics and p-values of Welch's T-test with hypothesis: score of honest and cheating participants have the same mean. Results suggest rejecting this hypothesis for most scenarios with high confidence, hence, there is a significant difference in the obtained quality scores between the two groups. }
    \label{tab:w-test}}
	\hfill
	\parbox{.48\linewidth}{
	\resizebox{0.48\textwidth}{!}{%
    \begin{tabular}{|cc|cccc|cccc|}
        \hline
         \multirow{2}{*}{Scenario} & \multirow{2}{*}{Type} & \multicolumn{4}{|c|}{Attacker} & \multicolumn{4}{c|}{Free-Rider} \\
         && MM & MC & CM & CC & MM & MC & CM & CC \\
         \hline
         \multirow{2}{*}{5/2/1} & p-value &2.2e-06 & 1.7e-15 & 1.3e-24 & 5.2e-34 & 1.0e-10 & 6.3e-20 & 1.9e-29 & 1.2e-38\\
         & Stat &324 & 2889 & 8021 & 15754 & 1278 & 5124 & 11577 & 20585\\
         \hline
         \multirow{2}{*}{25/5/1} & p-value &1.0e-04 & 1.1e-05 & 2.4e-06 & 9.8e-11 & 1.3e-04 & 6.0e-05 & 3.1e-07 & 6.5e-13\\
         & Stat &1682 & 12956 & 34047 & 70106 & 5864 & 21418 & 48965 & 92818\\
         \hline
         \multirow{2}{*}{25/5/2} & p-value &1.0e-06 & 3.7e-09 & 3.1e-12 & 4.5e-20 & 1.1e-03 & 1.0e-10 & 2.1e-13 & 7.6e-25\\
         & Stat &3115 & 24684 & 66824 & 135688 & 9732 & 43211 & 95119 & 180187\\
         \hline
         \multirow{2}{*}{100/10/1} & p-value &1.8e-01 &5.8e-04& 1.1e-03 &2.3e-04& 2.3e-02& 1.2e-02 &6.5e-04 &7.6e-04\\
         & Stat &4724 &49173& 126717 &246759 &20410 &78120 &180946 &312196 \\
         \hline
         \multirow{2}{*}{100/10/2} & p-value &2.3e-02& 3.0e-03& 1.8e-06& 4.1e-08& 1.2e-02& 2.1e-04& 2.8e-06 &2.2e-09\\
         & Stat &10107& 87000& 254909 &497224& 38762 &157764& 358577 &652865 \\
         \hline
         \multirow{2}{*}{100/10/3} & p-value &1.8e-04& 6.7e-04& 1.7e-04 &1.6e-09& 1.0e-02& 2.4e-03& 7.8e-06& 1.5e-11\\
         & Stat &16532& 128214& 347931& 723235& 55848& 218575& 508582 &954175\\
         \hline
         \multirow{2}{*}{100/10/4} & p-value &5.5e-04& 1.6e-06 &2.4e-09 &7.8e-14& 3.0e-04& 6.8e-07& 1.2e-10 &1.6e-14\\
         & Stat &20536& 176924& 488678& 970658& 77018 &307794 &701928& 1257416\\
         \hline
         \multirow{2}{*}{100/10/5} & p-value &1.9e-04 &6.2e-06& 1.0e-11 &7.4e-15 &4.5e-03& 3.0e-08& 4.4e-11& 1.1e-20\\
         & Stat &25268 &211680 &608156& 1184301& 89505& 381342 &854310& 1586432\\
        \hline
    \end{tabular}}
    \caption{Statistics and p-values of the Mann–Whitney U-test with hypothesis: the distributions of the score for honest and cheating participants are the same. Results suggest rejecting this hypothesis for most scenarios with high confidence, hence, there is a significant difference in the obtained quality scores between the two groups. }
    \label{tab:wm-test}}
\end{table*}

\begin{table*}[h]
    \centering
    \parbox{.48\linewidth}{
   	\resizebox{0.48\textwidth}{!}{%
    \begin{tabular}{|cc|cccc|cccc|}
        \hline
         \multirow{2}{*}{Scenario} & \multirow{2}{*}{Type} & \multicolumn{4}{|c|}{Attacker} & \multicolumn{4}{c|}{Free-Rider} \\
         && MM & MC & CM & CC & MM & MC & CM & CC \\
         \hline
         \multirow{2}{*}{5/2/1} & p-value &nan & 2.3e-03 & 3.2e-07 & 1.3e-04 & nan & 2.0e-07 & 1.9e-07 & 7.9e-21\\
         & Stat &nan & 25.6 & 47.5 & 33.0 & nan & 48.6 & 48.7 & 116.2\\
         \hline
         \multirow{2}{*}{25/5/1} & p-value &1.5e-01 & 1.5e-04 & 5.4e-05 & 1.0e-08 & 3.9e-03 & 6.8e-04 & 2.9e-06 & 1.7e-10\\
         & Stat &13.4 & 32.7 & 35.2 & 55.4 & 24.3 & 28.9 & 42.3 & 64.6\\
         \hline
         \multirow{2}{*}{25/5/2} & p-value &6.8e-06 & 1.9e-10 & 1.1e-14 & 6.0e-27 & 7.7e-03 & 7.5e-12 & 7.0e-17 & 6.1e-39\\
         & Stat &38.3 & 62.0 & 83.3 & 146.0 & 20.8 & 69.1 & 96.8 & 203.5\\
         \hline
         \multirow{2}{*}{100/10/1} & p-value &4.3e-01& 2.4e-03& 4.2e-02& 8.7e-02& 3.0e-01 &3.2e-02& 6.7e-02& 1.8e-01 \\
         & Stat &9.0& 25.6& 17.4& 15.2& 10.7& 18.3& 16.0& 12.7\\
         \hline
         \multirow{2}{*}{100/10/2} & p-value &4.5e-01 &1.9e-02& 4.7e-07 &1.1e-06& 1.7e-01& 2.9e-04& 4.2e-06 &1.5e-07\\
         & Stat &8.9& 19.8 &46.6 &44.6& 12.8& 31.1& 41.4& 49.3\\
         \hline
         \multirow{2}{*}{100/10/3} & p-value &1.0e-03 &7.0e-0& 1.2e-02& 1.7e-07 &7.2e-04& 1.4e-02& 1.5e-03& 2.3e-09\\
         & Stat &27.8& 28.8& 21.1& 49.0 &28.7& 20.7& 26.8& 58.8\\
         \hline
         \multirow{2}{*}{100/10/4} & p-value &1.0e-02& 9.4e-04& 4.9e-06 &1.1e-10& 2.1e-03& 1.3e-03& 4.5e-07 &7.5e-11\\
         & Stat &21.6 &28.0& 41.1& 65.5 &26.0 &27.3& 46.7& 66.4\\
         \hline
         \multirow{2}{*}{100/10/5} & p-value &1.9e-02& 1.4e-03& 2.1e-08& 9.1e-13 &2.6e-01& 2.0e-05& 3.7e-08 &5.3e-17\\
         & Stat &19.8& 26.9 &53.8& 76.2& 11.2& 37.6& 52.5& 97.4\\
        \hline
    \end{tabular}}
    \caption{Statistics and p-values of the Chi-Squared test (using 10 buckets) with hypothesis: the frequencies of the score for honest and cheating participants are the same. Results suggest rejecting this hypothesis for most scenarios with high confidence, hence, there is a significant difference in the obtained quality scores between the two groups. }
    \label{tab:cs-test}}
	\hfill
	\parbox{.48\linewidth}{
	\resizebox{0.48\textwidth}{!}{%
    \begin{tabular}{|cc|cccc|cccc|}
        \hline
         \multirow{2}{*}{Scenario} & \multirow{2}{*}{Type} & \multicolumn{4}{|c|}{Attacker} & \multicolumn{4}{c|}{Free-Rider} \\
         && MM & MC & CM & CC & MM & MC & CM & CC \\
         \hline
         \multirow{2}{*}{5/2/1} & p-value &2.3e-09 & 1.2e-20 & 8.9e-16 & 9.7e-50 & 1.0e-11 & 2.5e-27 & 6.7e-16 & 3.9e-59\\
         & Stat &1.0 & 0.9 & 0.9 & 0.9 & 0.9 & 0.9 & 0.9 & 0.9\\
         \hline
         \multirow{2}{*}{25/5/1} & p-value &1.4e-03 & 7.3e-05 & 4.8e-04 & 5.1e-08 & 1.3e-03 & 1.6e-03 & 6.1e-05 & 5.9e-10\\
         & Stat &0.6 & 0.4 & 0.3 & 0.4 & 0.4 & 0.3 & 0.3 & 0.4\\
         \hline
         \multirow{2}{*}{25/5/2} & p-value &1.2e-05 & 2.1e-08 & 1.3e-10 & 6.8e-17 & 1.6e-03 & 1.5e-09 & 1.0e-11 & 1.1e-16\\
         & Stat &0.6 & 0.4 & 0.4 & 0.4 & 0.3 & 0.4 & 0.4 & 0.4\\
         \hline
         \multirow{2}{*}{100/10/1} & p-value &4.2e-01& 3.8e-03& 1.0e-02& 4.0e-03& 1.2e-01& 7.5e-02& 3.8e-03 &7.4e-03 \\
         & Stat &0.28& 0.33 &0.24& 0.22& 0.27& 0.21&0.24 &0.20\\
         \hline
         \multirow{2}{*}{100/10/2} & p-value &3.0e-01& 1.9e-02 &2.5e-04& 5.9e-05 &1.4e-01 &1.2e-03 &4.2e-04& 3.3e-06\\
         & Stat &0.22& 0.21& 0.22& 0.20& 0.19& 0.23& 0.20 &0.22\\
         \hline
         \multirow{2}{*}{100/10/3} & p-value &1.0e-04 &7.8e-04& 8.0e-04& 2.9e-09& 1.8e-02 &1.8e-02& 1.0e-05& 5.6e-11\\
         & Stat &0.42& 0.22& 0.17& 0.23& 0.21& 0.15& 0.20& 0.24 \\
         \hline
         \multirow{2}{*}{100/10/4} & p-value &1.9e-03& 2.4e-04 &1.2e-06& 1.0e-09&4.0e-04& 1.1e-04& 1.7e-07 &6.2e-11\\
         & Stat &0.31& 0.21 &0.20 &0.21 &0.24& 0.19& 0.20 &0.21\\
         \hline
         \multirow{2}{*}{100/10/5} & p-value &8.6e-03& 1.2e-03& 1.7e-07& 1.1e-08 &9.0e-02& 2.0e-05 &3.6e-07 &4.8e-14\\
         & Stat &0.25 &0.17& 0.19& 0.18& 0.13& 0.18& 0.17& 0.21 \\
        \hline
    \end{tabular}}
    \caption{Statistics and p-values of the Kolmogorov-Smirnov test with hypothesis: the distributions of the score for honest and cheating participants are the same. Results suggest rejecting this hypothesis for most scenarios with high confidence, hence, there is a significant difference in the obtained quality scores between the two groups.}
    \label{tab:ks-test}}
\end{table*}

\end{document}